\documentclass[twocolumn,conference]{IEEEtran}
\usepackage[T1]{fontenc}
\usepackage[latin9]{inputenc}
\usepackage[unicode=true,
 bookmarks=true,bookmarksnumbered=true,bookmarksopen=true,bookmarksopenlevel=1,
 breaklinks=false,pdfborder={0 0 0},pdfborderstyle={},backref=false,colorlinks=false]
 {hyperref}
\hypersetup{pdftitle={Your Title},
 pdfauthor={Your Name},
 pdfpagelayout=OneColumn, pdfnewwindow=true, pdfstartview=XYZ, plainpages=false}

\makeatletter
\newcommand{\lyxaddress}[1]{
	\par {\raggedright #1
	\vspace{1.4em}
	\noindent\par}
}

\usepackage[caption=false,font=footnotesize]{subfig}

\makeatother

\begin{document}
\title{Practical Challenges in Landing a UAV on a Dynamic Target}
\author{\IEEEauthorblockN{Adarsh Salagame}\IEEEauthorblockA{Department of Aerospace\\
Indian Institute of Science\\
Bangalore\\
Email: adarshsalagame@gmail.com\\
}\and \IEEEauthorblockN{Sushant Govindraj}\IEEEauthorblockA{Department of Aerospace\\
Indian Institute of Science\\
Bangalore\\
Email: sushant.govindraj97@gmail.com}\and \IEEEauthorblockN{S. N. Omkar}\IEEEauthorblockA{Department of Aerospace\\
Indian Institute of Science\\
Bangalore\\
Email: omkar@iisc.ac.in}}
\maketitle

\lyxaddress{A white paper based on research conducted at the MAV Lab under Dr. S N Omkar, for the Wipro IISc Research and Innovation Network (WIRIN)\\
Copyright by authors.\\
}
\begin{abstract}
Unmanned Aerial Vehicles grow more popular by the day and applications for them are crossing boundaries of science and industry, with everything from aerial photography to package delivery to disaster management benefiting from the technology. But before they become commonplace, there are challenges to be solved to make them reliable and safe.

The following paper discusses the challenges associated with the precision landing of an Unmanned Aerial Vehicle, including methods for sensing and control and their merits and shortcomings for various applications.
\end{abstract}

\begin{IEEEkeywords}
unmanned aerial vehicles, autonomous flight, application, challenges,
control
\end{IEEEkeywords}

\section{Introduction}

Precision landing is a deceptively complex task, yet one with many applications. This complexity can be attributed to the many different variables that factor in based on the applications in which this technology is used. These applications can be broadly classified into four types based on the target upon which to land:
\begin{enumerate}
\item Ground-based landing on a static target
\item Ground-based landing on dynamic target
\item Sea-based landing
\item Air-based landing
\end{enumerate}
Due to the nonlinear nature of Sea and Air based systems, these targets have been considered dynamic. Applications for each of these is discussed below and the different methodologies to approach these problems, their key challenges, and shortcomings are discussed in Sections \ref{sec:Sensing-Methods} and \ref{sec:Control-Methods}.

\subsection{Ground-based landing on static target}

This includes landing on buildings, parked vehicles, or simply designated areas of the ground. Applications for Ground-based landing on a static target are package delivery for consumers, medical or disaster management use cases, and autonomous docking of a drone on a charging station, among others. Package delivery in particular is a popular area of interest, with companies like Google, Amazon, and Uber implementing urban solutions for rooftop or front lawn delivery, and companies like Zipline implementing the delivery of medical supplies to remote rural locations using drones carrying parachutes. With ground-based static landing, in most applications, precise landing is not a top priority, with the target size in the order of a few meters. However, this still poses other challenges like privacy and security, which will be discussed in Section \ref{sec:Other-Challenges}.

\subsection{Ground-based landing on a dynamic target}

Dynamic targets are generally vehicles such as cars and vans. Use cases for this technology would be surveillance, package delivery, and scouting. Google has experimented with using a ground vehicle in tandem with a drone for last-mile package delivery, and the process would be greatly optimized if the drone could be deployed and recollected while the ground vehicle was in motion. Similarly, scouting ahead of envoys for security is another application, where the drone takes off and lands on a moving target. In indoor environments, drones could be used for warehouse inventory management in tandem with an unmanned ground vehicle (UGV) as another use case for this technology.

In each of these use cases, the target is of varying levels of sophistication, from a simple van to a UGV, and the challenges posed in each are different. Landing on a dynamic platform typically occurs in three stages. 
\begin{enumerate}
\item \emph{Approach}: The drone must find the target vehicle and lock onto it using some sensing system such as a camera or radio frequency.
\item \emph{Tracking}: The drone must track and follow above the target vehicle using some control algorithm, maintaining a position above the landing pad.
\item \emph{Descent}: While continuing to track the target, the drone must reduce altitude until it can safely land and disarm.
\end{enumerate}
Each step poses unique challenges. In the approach stage, the target vehicle must successfully be identified. For a car or a van, this means differentiating it from other vehicles present on the road. Once identified, the drone must reduce the distance to the vehicle and begin tracking. Aggressive maneuvers at this stage can lead to overshooting and losing the lock on the target. Over-aggressive maneuvers also pose a safety risk to any nearby people or objects. On the other hand, under aggressive maneuvers can mean the drone is unable to catch up with and land on the target within a reasonable amount of time. The system must therefore be robust and adaptable, adjusting to the speed of the target, while maintaining safe movements. In the tracking stage, in order to consistently track the target, its motion must be estimated, modeled, and predicted and the non-linearities must be factored in. This requires an optimized control algorithm driven by precise state estimation of the target. In addition, the tracking stage poses similar challenges with regard to aggressiveness, with over-aggressive maneuvers quickly leading to uncontrolled oscillations of the drone.

Finally, in the descent stage, the drone reduces its altitude for landing. As the altitude lowers, the allowable error in tracking reduces, and control must be highly precise. For vision-based systems, this is made more difficult by the reduced field of view for the camera. The effects of wind and the ground effects from the propellers must also be countered, which becomes more significant, the closer the drone is to the target. As the drone touches down on the target, mechanisms must be in place to safely secure the drone to the landing pad to prevent it from flipping over or slipping off in motion.

\subsection{Sea based landing}

This is a very specific use case, for landing on ships or boats. The challenges for this are similar to that of landing on a ground-based dynamic target, with the added complexity, however, of a continuous pitch and roll caused by the motion of the sea. To compensate for this, the adaptive landing gear can be used, which is discussed in Section \ref{sec:Landing}. Applications for this technology is search and rescue missions by coast guards using drones to aid in surveillance and scouting and mapping for sea-based navigation.

\subsection{Air based landing}

Air-based landing is air-to-air docking, which can be used for mid-air refueling. \cite{Chen2015} has used a large drone equipped with multiple smaller drones used as flying batteries to extend the flight time significantly. Such applications are subject not only to the challenges of ground and sea-based landing but face additional challenges due to backdraft from the propellers of both aerial vehicles and having more degrees of freedom for noise.

\section{Sensing Methods\label{sec:Sensing-Methods}}

The approach stage depends heavily on the method of sensing used to identify the target. Different methods are suitable for different applications and all involve a trade-off between accuracy and speed. The different possible methods of sensing are listed below.

\subsection{GPS Based Way-point Navigation}

Way-point navigation is a popular standard for autonomous flight. The biggest merit of this approach is the simplicity and minimal computation required. A list of set points in the form of latitude, longitude, and altitude is set before the flight and the drone autonomously follows these set points. Without the need for any external computation, additional hardware such as micro-controllers, which not only add weight to the drone but also draw a significant amount of power and reduce flight time, can be done away with, relying purely on the flight controller. GPS-based navigation is useful in cases where the expected path of the target is known or the target is advertising its current GPS location and the precision of landing is not a priority. However, due to the large error in measurement (2-5m) of GPS, this cannot be used in cases requiring high precision for landing. Additionally, if the target is passive, having no communication with the drone, the drone would be landing blind, with no checks on the accuracy of the landing. This method also fails in GPS-denied environments such as remote or covered areas. An alternative for high precision high accuracy GPS navigation is the Real-time Kinematic (RTK) GPS, which has an error of just 1cm and a precision of 0.1mm with negligible latency. However, RTK requires a constant communication link between a GNSS/GPS base and the drone. In case this communication link is lost RTK will not work anymore. This base station is bulky, must not be moved, and requires large amounts of time for initial calibration. This is, therefore not a practical solution for this application. RTK GPS is used primarily in surveying applications where a high degree of precision and accuracy is required within a fixed area. Another general algorithm for a GPS-based tracking and landing system would be to plan a trajectory from the current location to the target destination using the bearing of the drone which is the angle between the North and the direction to the destination. The system needs to apply a velocity in the body coordinate system so the bearing would have to be compared to the heading of the drone prior to that. The set of equations governing the velocity of the drone in the x and y directions uses the yaw and the heading velocity of the system as a whole. The computation resulting from the recalculation of the trajectory between the current position and the target position is not hefty. The system needs to iteratively update the distance between the current coordinate and the target coordinate, so as to slow the drone down while reaching the landing platform. Their proposed method uses position covariances provided by flight controller software which is sent together with the current geodetic coordinate of the drone. Unscented Transform is used to calculate bearing between 2 coordinates considering those 2 coordinates have covariances. The problem lies in the position covariance provided by the Autopilot Software, which is in meters. One could use Kalman Filters to tune the measurement of position variance Navigation using the course-over-ground algorithm is not more reliable than the navigation algorithm with GNSS and Compass at a navigation distance of less than 1 m. This tracking algorithm does have its weaknesses, it is not good for short-distance navigation, so when the drone is nearing the goal position, there will be some difficulty reaching the landing. The GNSS module is error prone which is why this problem occurs. Windy conditions also affect the drone, and as a result of the flight trajectory never matches the planned trajectory, but the proposed algorithm will ensure the drone will approach and land in the goal position.

\subsection{Radio Frequency Based Navigation}

Radiofrequency communication-based navigation techniques usually entail using a beacon or transmitter on the moving target, enabling a two-way communication between the drone and the dynamic object. \cite{4341511} are capable of measuring the distance between two radio devices, and this distance measuring technique coupled with the use of messages carrying landing information such as velocity and heading direction can guide the drone to land on the target while in motion. However, this method is constrained by such applications that require the target to be as passive and as adaptable as possible. Military applications requiring radio silence or consumer applications where the target is a regular vehicle fall in this category. Moreover, the range of communication between the drone and the target is limited in areas where line of sight communication is not possible. Commercial aircraft use a non-directional radio beacon for landing, however, this is subject to a plethora of factors that can cause noise in the system, including terrain effect - where high-altitude cliffs or mountain ranges due to their magnetic deposits can erode transmissions, station interference - caused by congestion in the low-frequency (LF) and mid-frequency bands (MF) and dip angle - caused by the pitching and rolling of drones causing erroneous readings for the receiver. 

\subsection{Vision Based Navigation}

Vision is a staple of any autonomous system because of its versatility. Using just camera frames, a large amount of information can be extracted to provide scene understanding, and localization and assist in navigation. Based on the application, different types of cameras can be selected. Fish eye lenses provide a large field of view, but require extra computation to correct distortion in each frame. This also leads to a loss of resolution in the image. Monocular cameras have a limited frame, but images can have a greater resolution as a result. Stereo cameras offer a slightly wider field of view than monocular and also provide additional information such as depth. However, these are computationally heavy and the need for synchronization of the cameras adds additional complexity. The applications listed below detail use cases for these cameras and the different vision-based sensing techniques that can be implemented.

\subsubsection{Optical Flow}

Optical Flow techniques are quite powerful and used in multifarious applications. It can be either dense or sparse optical flow depending on the number of feature points tracked. Once a set or locus of pixels are tracked, the algorithm gives the displacement in the x and y spatial directions so as to make it convenient to track the movement of a given object in frame. In this case, the object most likely should be the moving platform upon which the drone would land. This technique needs to be paired with basic image segmentation techniques such as Scale-Invariant Feature Transform (SIFT), Semantic segmentation, or Dilation and Erosion. Optical Flow can also be paired with more complex deep learning models in order to increase accuracy such as YOLO, R-CNN, fast R-CNN, faster R-cNN and Mask R-cNN.

\subsubsection{Marker Based Landing}

With Optical Flow and Machine learning based methods, the drone can theoretically track and land on any target without any prior preparation. However, this is paid for in terms of computation, with both these methods being computationally intense. A simpler solution is to use visual markers to differentiate the target for the drone's camera. These markers can also provide information regarding the relative position and the altitude of the drone with respect to the target if the size of the marker is known prior to flight. \cite{8088164} uses a simple cross as a marker, using image processing techniques to identify and localize with respect to it.  \cite{Araar2017,Ling} use fiducial markers for the same purpose, reducing computation. Marker based tracking and landing is very accurate with minimal latency and moderate computation, making it a good methodology to adopt, with companies like Google and Amazon working ArUco marker based landing platforms for package delivery. However, it is subject to environmental and logistical constraints. Poor lighting conditions and occlusion due to dirt or obstacles can affect the visibility of the markers, leading to difficulty or inability to land. In addition, the landing platform must be prepared specially with markers of a specific known size, and it must be regularly maintained to ensure the markers are visible and in high contrast with the background. Moreover, the platform must be lit up in low light conditions, adding to the cost and complexity of the solution.

\subsubsection{Infrared Based Landing}

An alternative to using printed or painted markers would be to use Infrared lights arranged in a pattern to convey similar information. \cite{Wenzel2011} uses a Wii remote and a camera to successfully track and land on a moving platform in indoor conditions. These require very little power and work well in low light conditions. However, infrared detection performs poorly in direct sunlight, making this impractical as a stand alone solution. Coupling this with printed or painted markers on the landing platform could potentially provide a more well rounded solution, although the problem of occlusion and maintenance still persists. 

\subsubsection{Kalman Filtering}

Although this isn't strictly a sensing method, this belongs on this list for it\textquoteright s significant use in sensing. Kalman Filtering is a method of state estimation that corrects for measurement and system noise to increase accuracy. It can also be used to estimate additional in-formation about a system that is not directly measured by fusing sensor readings from other related state quantities. This relies on having models for the state transition and measurement of the system variables, and the assumption that all system and measurement noise is Gaussian in nature. Kalman filtering can not only increase the accuracy of state estimation from noisy sensor measurements but can also reduce the amount of variation in state estimation, making the system more stable. The classic Kalman Filter works well for single variable linear applications but performs poorly when non-linearities and cross-correlation of states are involved. Variants of this such as the Extended Kalman Filter (EKF) and the Unscented Kalman Filter (UKF) provide better results in such situations and are used in most real-world applications.

\section{Control Methods\label{sec:Control-Methods}}

The sensing system present on the drone outputs a relative position of the target with respect to the drone. This is used in the second stage, which is tracking. The information from the sensing system is input into a control algorithm that adjusts the motion of the drone to perform maneuvers to catch up and follow the target. This control can be in the form of velocity, position set points, or attitude. Position set points rely on accuracy of GPS, and are therefore not a very reliable control method. Attitude and velocity are inter-convertible given the dynamics model of the drone, and can therefore be considered together, although while attitude control provides more precision, velocity control is more intuitive and simple. In cases where the model is not available, velocity can be used as a reliable control method. The most important thing with any control algorithm is the stability of the control. The drone must settle into the following motion fast, with minimal oscillation about the target. The different control algorithms that can be implemented are discussed below.

\subsection{PID Controllers}

Proportional Integral Derivative Controllers (PID) are the most popular method for converging onto a target, due to their simplicity and ease of implementation, requiring only three gains to be set, which directly affect the rate of convergence and damping of the system. Over the years, for various applications, the PID controller has been modified and customized to improve results based on specific use cases. Three of the prominent variants of the PID Controller are discussed below.

\subsubsection{Classical PID}

The original form of the PID controller is still popular and in use today, largely due to its simplicity. Requiring minimal computation and with the gains fixed, the response is predictable and controllable once tuned. However, the biggest disadvantage of the classical PID is the difficulty in tuning the Proportional, Derivative, and Integral gains. There are certain methods such as the Ziegler-Nichols method and Cohen and Coon method for closed loop and open loop tuning, however, without accurate real-world models of the system, these methods cannot be reliably used to tune a PID. Moreover, even if the controller is well-tuned, it performs poorly when the system is highly non-linear and is not robust enough to track a varying range of target speeds and heading directions.

\subsubsection{Fuzzy PID}

An established variation over the classical PID is the Fuzzy Controller. \cite{7979027} successfully uses a Fuzzy PID for tracking and landing on a target at speeds of up to 10 m/s. In this, the gains are set dynamically within a given range based on a fuzzy rule-base. This allows the system to adapt to any non-linearities, giving significantly more accurate results than the classical PID. This, however, depends entirely on the quality of the fuzzy rule-base upon which the gains are set. Creating a good rule-base requires a deep knowledge of the field and a good understanding of system responses. This can be a severe limitation to scaling applications to various conditions and use cases, as expertise is required in each area to implement this well.

\subsubsection{Model Predictive Control}

Model predictive Control (MPC) is a multivariate control technique that offers a wider range of tune-able parameters. It can be used to make predictions about the system's future behavior. It helps solve optimization algorithms to find the best control output that drives the predicted output to the reference. MPC can handle Multi-Input Multi-Output (MIMO) systems that can have multiple interactions between their inputs and outputs. It offers many features that help improve classical PID controller performance. This series also discusses MPC design parameters such as the controller sample time, prediction and control horizons, constraints, and weights. It also gives you recommendations for choosing these parameters. You\textquoteright ll learn about adaptive, gain-scheduled, and non- linear MPCs, and youâ\u{A}\'{Z}ll get implementation tips to reduce the computational complexity of MPC and run it faster. There are essentially two kinds of Model Predictive Control, they are Non-linear Model Predictive Control model (NMPC) and classical linear model predictive control (LMPC). NMPC is a much more advanced method than the classical LMPC as it takes full system dynamics into consideration which, also presents to be more complex. \cite{6805732} shows that the x, and y errors in maintaining the position set-points is much lesser in NMPC and when it is subjected to wind and or external disturbances then the errors are still more pronounced only with respect to the LMPC model. The thrust command's response and the step response in the x-direction is faster for NMPC than LMPC without an overshoot in due to it's full usage of the system's dynamics. An aggressive trajectory tracking in high speed winds using higher-order position equations was experimented with. Computationally, NMPC is much less time consuming than LMPC which makes it very convenient. It makes sense to use the NMPC algorithm for tracking and landing the drone without any additional time consumption and computation. It also offers a better accuracy rate when subjected to external disturbances as well. MPC offers a better PID control in case there are multiple PID controllers in the system, fuzzy controllers on the other hand are still quite inefficient when it comes to handling multiple PIDs because of the increase in computation and complexity of the algorithm.

\subsection{LQR Controller}

This work has been done in an indoor environment using opti-track cameras and ground stations as aides to perform the take-off, tracking and landing tasks. The UAV's controller is based on a combination of Sliding Mode Control (SMC) and Linear Quadratic Regulator (LQR). The Formation controller based on SMC is used to allow the UAV to track the Unmanned Ground Vehicle (UGV) in the leader-follower manner. The proposed algorithms are capable of allowing the UAV to take-off, track and land on the moving UGV. The main advantage of the proposed controller is its ability to apply in real-time as shown in the experimental results. The controller succeeds to perform the mission requirements in two different scenarios. Furthermore, it is illustrated that the control system is stabilized with good performance for accomplishing the specified missions.

\subsection{Deep Learning}

Machine learning is one of the most versatile and intelligent techniques, to perform either prediction or classification. This technique if developed in the right way yields a diminished loss and high accuracy. However, such systems no matter how intelligent have trade-offs that include, high resources such as GPU training time, latency in model development and prototyping. When it comes to deployment, it still requires the usage of a good GPU and does have higher latency issues compared to hard computing techniques. Having a higher number of hidden layers and nodes surely has a negative effect on time consumption by the system. Drone based applications make it all the more difficult to deploy such Machine learning algorithms since the system needs to be light-weight, low power and extremely fast. In \cite{Rodriguez-Ramos2019}, Deep Reinforcement Learning (DRL) is used to realize a system for the autonomous landing of a quadrotor on a static target. The overall performances are comparable with an AR-tracker algorithm and manual human piloting. The system as a whole is faster autonomously than humans (manually controlling the drone) in reaching the target and is more immune to physical intrusions of the marker compared to the AR-tracker. The network makes it very convenient by generalizing the environmental features and performing well even though it is trained on a minimal subset of textures. When it comes to the missed landings, the flight was only interrupted by the expiration time, which controlled the amount of time spent on air. The drone found itself to be landing within the target region at all times. Almost all of the missed landing have been caused by extreme external environment conditions (mutable lighting and strong drift), which were not modeled in the simulator. The outputs can be further improved by taking into account the factors during the training phase. The results obtained are good but still need lots of areas to improve upon for real-world deployment.

\section{Landing\label{sec:Landing}}

The final stage is descent. In order to land safely on a dynamic platform, there must be mechanisms or systems in place to ensure the drone maintains it\textquoteright s position on the landing platform. In applications such as landing on the deck of a ship, there is not only a linear motion, but also an element of roll and pitch to the landing surface. Similar challenges are also faced in applications such as landing on a ground vehicle that is moving on an uneven surface, or docking with an aerial vehicle. The following are possible solutions to these challenges.

\subsection{Adaptive Landing Gear}

The fixed landing gear on drones gives them a tendency to bounce or flip over when landing at an angle to the surface. Adaptive landing gear can solve this problem by adapting to the uneven motions of the surface. There are different types of adapting landing gear. One can use linear actuators to match each leg to the level of the surface independently while the body of the drone remains stable. \cite{8291148} uses landing gear with a degree of freedom, allowing them to bend to match the angle and unevenness of the surface. These solutions work well to adapt to the pitch and roll of the landing surface, but do little to prevent slippage from the surface once landed. Changes in the momentum of the moving platform would dislodge the landed drone, making this only a partial solution in most practical cases.

\subsection{Magnetic Landing Gear}

To secure the drone once it has landed, magnetic landing gear is a potential solution. This would require heavier metallic platforms and a more sophisticated landing gear mechanism, but it would ensure that the drone is secured once it has landed.

\section{Other Challenges\label{sec:Other-Challenges}}

Since the drones that require precision landing measures, are the drones that have applications mostly for drone based package delivery, security and safety measures would narrow down to only the said application.

\subsection{Privacy, Security and Safety Challenges}

Most precision landing mechanisms being developed today use some form of machine intelligence and cameras, which are constantly collecting data for better performance. This flying camera-in-the-sky has deep privacy implications, with drones potentially flying over populated residential and commercial areas and capturing private scenes. In a world driven by data, devices such as these, no matter how benign can pose a large risk to safety.

\subsection{Legal Challenges}

The legal issues are faced by drones intentionally or unintentionally not conforming to the laws and regulations of a certain society. For example, India's national aviation authority, the Ministry of Civil Aviation, states that flying a drone is legal in India, but not by foreigners. All drones need to have a Unique Identification Number (UIN). Permits are needed for flying commercially and drone pilots are to always maintain a Line of Sight (LOS) while flying the drone. A height of 400 feet is the maximum. In this case, package delivery and similar autonomous delivery applications cannot be feasible due to this pilot LOS rule and height limit. There are other regulations that do not allow drones to fly over private homes. This would complicate the trajectory planning when the drone has to deliver a package in a house within a populated neighborhood, in this case, there would not be any other option other than to fly over the house to get the job done. There are also legal gray areas where the drones may land by accident on another individual\textquoteright s property. The drone\textquoteright s navigational algorithm must be very accurate to the centimeter and the battery before any flight must be checked so as to avoid emergency automatic landing on private property. UAVs using vision-based navigation techniques face a dilemma. Private properties are not supposed to be photographed by non-governmental organizations or Corporate entities and vision-based navigation entails using downward and forward-facing cameras to capture features, markers, and/or landing platforms, some of the background information including these private properties would be captured in the landing phase.

\subsection{Environmental Challenges}

Drones are notorious when it comes to noise production. There are certain societies and/or places that ban the production of noise above a specific decibel value. In this situation, drone industries face the responsibility of creating drones with reduced noisy elements, or flying them at higher altitudes for package delivery, both of which are quite challenging. Another challenge is to deploy drones that do not physically harm birds or planes flying in the air, or trees, houses, and people (free-falling drones in a populated area). A solution is to use propeller guards and lighter drones that would not harm any flying birds or humans. In addition, long-range drones often use both petrol and battery as power sources to maximize endurance, which increases the carbon footprint.

\section{Conclusion}

This paper has summarized and discussed some of the most popular methods of landing an Unmanned Aerial Vehicle on a moving platform, talking about the challenges faced with each of these. It also includes some general challenges with bringing such a system into the consumer space, including legal and environmental challenges. Alternatives and potential solutions have also been mentioned, but this is by no means comprehensive. As is always the case with engineering, these systems are continuously evolving, and newer solutions are constantly being developed. This paper endeavors to cover the most common approaches to this problem.

\bibliographystyle{IEEEtran}
\nocite{*}
\bibliography{bibliography}

\begin{thebibliography}{10}
\providecommand{\url}[1]{#1}
\csname url@samestyle\endcsname
\providecommand{\newblock}{\relax}
\providecommand{\bibinfo}[2]{#2}
\providecommand{\BIBentrySTDinterwordspacing}{\spaceskip=0pt\relax}
\providecommand{\BIBentryALTinterwordstretchfactor}{4}
\providecommand{\BIBentryALTinterwordspacing}{\spaceskip=\fontdimen2\font plus
\BIBentryALTinterwordstretchfactor\fontdimen3\font minus
  \fontdimen4\font\relax}
\providecommand{\BIBforeignlanguage}[2]{{%
\expandafter\ifx\csname l@#1\endcsname\relax
\typeout{** WARNING: IEEEtran.bst: No hyphenation pattern has been}%
\typeout{** loaded for the language `#1'. Using the pattern for}%
\typeout{** the default language instead.}%
\else
\language=\csname l@#1\endcsname
\fi
#2}}
\providecommand{\BIBdecl}{\relax}
\BIBdecl

\bibitem{Chen2015}
\BIBentryALTinterwordspacing
C.-I. Chen, R.~Koseluk, C.~Buchanan, A.~Duerner, B.~Jeppesen, and H.~Laux,
  ``Autonomous aerial refueling ground test demonstration--a
  sensor-in-the-loop, non-tracking method,'' \emph{Sensors (Basel,
  Switzerland)}, vol.~15, no.~5, pp. 10\,948--10\,972, May 2015,
  25970254[pmid]. [Online]. Available:
  \url{https://www.ncbi.nlm.nih.gov/pubmed/25970254}
\BIBentrySTDinterwordspacing

\bibitem{4341511}
A.~{Awad}, T.~{Frunzke}, and F.~{Dressler}, ``Adaptive distance estimation and
  localization in wsn using rssi measures,'' in \emph{10th Euromicro Conference
  on Digital System Design Architectures, Methods and Tools (DSD 2007)}, Aug
  2007, pp. 471--478.

\bibitem{8088164}
D.~{Falanga}, A.~{Zanchettin}, A.~{Simovic}, J.~{Delmerico}, and
  D.~{Scaramuzza}, ``Vision-based autonomous quadrotor landing on a moving
  platform,'' in \emph{2017 IEEE International Symposium on Safety, Security
  and Rescue Robotics (SSRR)}, Oct 2017, pp. 200--207.

\bibitem{Araar2017}
\BIBentryALTinterwordspacing
O.~Araar, N.~Aouf, and I.~Vitanov, ``Vision based autonomous landing of
  multirotor uav on moving platform,'' \emph{Journal of Intelligent {\&}
  Robotic Systems}, vol.~85, no.~2, pp. 369--384, Feb 2017. [Online].
  Available: \url{https://doi.org/10.1007/s10846-016-0399-z}
\BIBentrySTDinterwordspacing

\bibitem{Wenzel2011}
\BIBentryALTinterwordspacing
K.~E. Wenzel, A.~Masselli, and A.~Zell, ``Automatic take off, tracking and
  landing of a miniature uav on a moving carrier vehicle,'' \emph{Journal of
  Intelligent \& Robotic Systems}, vol.~61, no.~1, pp. 221--238, 2011.
  [Online]. Available: \url{https://doi.org/10.1007/s10846-010-9473-0}
\BIBentrySTDinterwordspacing

\bibitem{7979027}
W.~{Si}, H.~{She}, and Z.~{Wang}, ``Fuzzy pid controller for uav tracking
  moving target,'' in \emph{2017 29th Chinese Control And Decision Conference
  (CCDC)}, May 2017, pp. 3023--3027.

\bibitem{6805732}
M.~H. {Tanveer}, D.~{Hazry}, S.~F. {Ahmed}, M.~K. {Joyo}, F.~A. {Warsi},
  H.~{Kamaruddin}, Z.~M. {Razlan}, K.~{Wan}, and A.~B. {Shahriman}, ``Nmpc-pid
  based control structure design for avoiding uncertainties in attitude and
  altitude tracking control of quad-rotor (uav),'' in \emph{2014 IEEE 10th
  International Colloquium on Signal Processing and its Applications}, March
  2014, pp. 117--122.

\bibitem{Rodriguez-Ramos2019}
\BIBentryALTinterwordspacing
A.~Rodriguez-Ramos, C.~Sampedro, H.~Bavle, P.~de~la Puente, and P.~Campoy, ``A
  deep reinforcement learning strategy for uav autonomous landing on a moving
  platform,'' \emph{Journal of Intelligent \& Robotic Systems}, vol.~93, no.~1,
  pp. 351--366, 2019. [Online]. Available:
  \url{https://doi.org/10.1007/s10846-018-0891-8}
\BIBentrySTDinterwordspacing

\bibitem{8291148}
Y.~S. {Sarkisov}, G.~A. {Yashin}, E.~V. {Tsykunov}, and D.~{Tsetserukou},
  ``Dronegear: A novel robotic landing gear with embedded optical torque
  sensors for safe multicopter landing on an uneven surface,'' \emph{IEEE
  Robotics and Automation Letters}, vol.~3, no.~3, pp. 1912--1917, July 2018.

\bibitem{8598647}
C.~{Kyrkou}, S.~{Timotheou}, P.~{Kolios}, T.~{Theocharides}, and
  C.~{Panayiotou}, ``Drones: Augmenting our quality of life,'' \emph{IEEE
  Potentials}, vol.~38, no.~1, pp. 30--36, Jan 2019.

\bibitem{Puri2005ASO}
A.~Puri, ``A survey of unmanned aerial vehicles ( uav ) for traffic
  surveillance,'' 2005.

\bibitem{7554984}
S.~{Jin}, J.~{Zhang}, L.~{Shen}, and T.~{Li}, ``On-board vision autonomous
  landing techniques for quadrotor: A survey,'' in \emph{2016 35th Chinese
  Control Conference (CCC)}, July 2016, pp. 10\,284--10\,289.

\bibitem{Vergouw2016}
\BIBentryALTinterwordspacing
B.~Vergouw, H.~Nagel, G.~Bondt, and B.~Custers, \emph{Drone Technology: Types,
  Payloads, Applications, Frequency Spectrum Issues and Future
  Developments}.\hskip 1em plus 0.5em minus 0.4em\relax The Hague: T.M.C. Asser
  Press, 2016, pp. 21--45. [Online]. Available:
  \url{https://doi.org/10.1007/978-94-6265-132-6_2}
\BIBentrySTDinterwordspacing

\bibitem{1013656}
S.~{Saripalli}, J.~F. {Montgomery}, and G.~S. {Sukhatme}, ``Vision-based
  autonomous landing of an unmanned aerial vehicle,'' in \emph{Proceedings 2002
  IEEE International Conference on Robotics and Automation (Cat.
  No.02CH37292)}, vol.~3, May 2002, pp. 2799--2804 vol.3.

\bibitem{7577060}
E.~{Vattapparamban}, I.~{GÃŒvenÃ§}, A.~Ä. {Yurekli}, K.~{Akkaya}, and
  S.~{UluagaÃ§}, ``Drones for smart cities: Issues in cybersecurity, privacy,
  and public safety,'' in \emph{2016 International Wireless Communications and
  Mobile Computing Conference (IWCMC)}, Sep. 2016, pp. 216--221.

\bibitem{6224828}
D.~{Lee}, T.~{Ryan}, and H.~J. {Kim}, ``Autonomous landing of a vtol uav on a
  moving platform using image-based visual servoing,'' in \emph{2012 IEEE
  International Conference on Robotics and Automation}, May 2012, pp. 971--976.

\bibitem{7502574}
H.~{Lee}, S.~{Jung}, and D.~H. {Shim}, ``Vision-based uav landing on the moving
  vehicle,'' in \emph{2016 International Conference on Unmanned Aircraft
  Systems (ICUAS)}, June 2016, pp. 1--7.

\bibitem{6196930}
S.~{Gupte}, {Paul Infant Teenu Mohandas}, and J.~M. {Conrad}, ``A survey of
  quadrotor unmanned aerial vehicles,'' in \emph{2012 Proceedings of IEEE
  Southeastcon}, March 2012, pp. 1--6.

\bibitem{doi:10.1002/net.21818}
\BIBentryALTinterwordspacing
A.~Otto, N.~Agatz, J.~Campbell, B.~Golden, and E.~Pesch, ``Optimization
  approaches for civil applications of unmanned aerial vehicles (uavs) or
  aerial drones: A survey,'' \emph{Networks}, vol.~72, no.~4, pp. 411--458,
  2018. [Online]. Available:
  \url{https://onlinelibrary.wiley.com/doi/abs/10.1002/net.21818}
\BIBentrySTDinterwordspacing

\bibitem{6997750}
W.~{Kong}, D.~{Zhou}, D.~{Zhang}, and J.~{Zhang}, ``Vision-based autonomous
  landing system for unmanned aerial vehicle: A survey,'' in \emph{2014
  International Conference on Multisensor Fusion and Information Integration
  for Intelligent Systems (MFI)}, Sep. 2014, pp. 1--8.

\bibitem{5508131}
Y.~{Liu} and Q.~{Dai}, ``A survey of computer vision applied in aerial robotic
  vehicles,'' in \emph{2010 International Conference on Optics, Photonics and
  Energy Engineering (OPEE)}, vol.~1, May 2010, pp. 277--280.

\bibitem{6842377}
A.~{Gautam}, P.~B. {Sujit}, and S.~{Saripalli}, ``A survey of autonomous
  landing techniques for uavs,'' in \emph{2014 International Conference on
  Unmanned Aircraft Systems (ICUAS)}, May 2014, pp. 1210--1218.

\bibitem{6842381}
J.~{Kim}, Y.~{Jung}, D.~{Lee}, and D.~H. {Shim}, ``Outdoor autonomous landing
  on a moving platform for quadrotors using an omnidirectional camera,'' in
  \emph{2014 International Conference on Unmanned Aircraft Systems (ICUAS)},
  May 2014, pp. 1243--1252.

\bibitem{7580657}
P.~{Serra}, R.~{Cunha}, T.~{Hamel}, D.~{Cabecinhas}, and C.~{Silvestre},
  ``Landing of a quadrotor on a moving target using dynamic image-based visual
  servo control,'' \emph{IEEE Transactions on Robotics}, vol.~32, no.~6, pp.
  1524--1535, Dec 2016.

\bibitem{Cesetti2009}
\BIBentryALTinterwordspacing
A.~Cesetti, E.~Frontoni, A.~Mancini, P.~Zingaretti, and S.~Longhi, ``A
  vision-based guidance system for uav navigation and safe landing using
  natural landmarks,'' \emph{Journal of Intelligent and Robotic Systems},
  vol.~57, no.~1, p. 233, Oct 2009. [Online]. Available:
  \url{https://doi.org/10.1007/s10846-009-9373-3}
\BIBentrySTDinterwordspacing

\bibitem{6017133}
B.~{HerissÃ©}, T.~{Hamel}, R.~{Mahony}, and F.~{Russotto}, ``Landing a vtol
  unmanned aerial vehicle on a moving platform using optical flow,'' \emph{IEEE
  Transactions on Robotics}, vol.~28, no.~1, pp. 77--89, Feb 2012.

\bibitem{LingKevin2014}
\BIBentryALTinterwordspacing
{Ling, Kevin}, ``Precision landing of a quadrotor uav on a moving target using
  low-cost sensors,'' 2014. [Online]. Available:
  \url{http://hdl.handle.net/10012/8803}
\BIBentrySTDinterwordspacing

\bibitem{1206795}
S.~{Saripalli}, J.~F. {Montgomery}, and G.~S. {Sukhatme}, ``Visually guided
  landing of an unmanned aerial vehicle,'' \emph{IEEE Transactions on Robotics
  and Automation}, vol.~19, no.~3, pp. 371--380, June 2003.

\bibitem{7505370}
X.~{Chen}, S.~K. {Phang}, M.~{Shan}, and B.~M. {Chen}, ``System integration of
  a vision-guided uav for autonomous landing on moving platform,'' in
  \emph{2016 12th IEEE International Conference on Control and Automation
  (ICCA)}, June 2016, pp. 761--766.

\bibitem{5584396}
M.~A. {Olivares-Mendez}, I.~F. {Mondragon}, P.~{Campoy}, and C.~{Martinez},
  ``Fuzzy controller for uav-landing task using 3d-position visual
  estimation,'' in \emph{International Conference on Fuzzy Systems}, July 2010,
  pp. 1--8.

\bibitem{GRANCHAROVA2012254}
\BIBentryALTinterwordspacing
A.~Grancharova, E.~I. GrÃžtli, and T.~A. Johansen, ``Distributed mpc-based path
  planning for uavs under radio communication path loss constraints,''
  \emph{IFAC Proceedings Volumes}, vol.~45, no.~4, pp. 254 -- 259, 2012, 1st
  IFAC Conference on Embedded Systems, Computational Intelligence and
  Telematics in Control. [Online]. Available:
  \url{http://www.sciencedirect.com/science/article/pii/S1474667015404756}
\BIBentrySTDinterwordspacing

\bibitem{6564973}
Z.~{He} and J.~{Xu}, ``Moving target tracking by uavs in an urban area,'' in
  \emph{2013 10th IEEE International Conference on Control and Automation
  (ICCA)}, June 2013, pp. 1933--1938.

\bibitem{polvara2017autonomous}
R.~Polvara, M.~Patacchiola, S.~Sharma, J.~Wan, A.~Manning, R.~Sutton, and
  A.~Cangelosi, ``Autonomous quadrotor landing using deep reinforcement
  learning,'' 2017.

\end{thebibliography}

\end{document}